\documentclass[10pt,conference]{IEEEtran}
\usepackage{cite}
\usepackage{amsmath,amssymb,amsfonts}
\usepackage{algorithmic}
\usepackage{graphicx}
\usepackage{textcomp}
\usepackage{xcolor}

\usepackage{wrapfig}

\usepackage{multirow}
\usepackage[ruled,vlined]{algorithm2e}
\usepackage{hyperref}
\usepackage[english]{babel}
\usepackage{todonotes}
\usepackage{enumerate}
\usepackage{enumitem}
\usepackage{comment}
\usepackage{ragged2e}
\usepackage{soul}
\usepackage{caption}
\usepackage{booktabs}

\def\BibTeX{{\rm B\kern-.05em{\sc i\kern-.025em b}\kern-.08em
    T\kern-.1667em\lower.7ex\hbox{E}\kern-.125emX}}

\newcommand{\paper}{\textit{RhythmEdge }}

\begin{document}

\title{Demo: RhythmEdge: Enabling Contactless Heart Rate Estimation on the Edge}

\author{\IEEEauthorblockN{Zahid Hasan$^{\dagger}$, Emon Dey$^{\dagger}$, Sreenivasan Ramasamy Ramamurthy$^{\dagger}$, Nirmalya Roy$^{\dagger}$, Archan Misra$^\S$}
\IEEEauthorblockA{\textit{$^\dagger$Mobile Pervasive \& Sensor Computing Lab, Center for Real-time Distributed Sensing and Autonomy (CARDS)}\\ \textit{$^\dagger$Department of Information Systems, University of Maryland, Baltimore County, United States} \\
\textit{$^\S$School of Computing \& Information Systems, Singapore Management University, Singapore} \\
\{zhasan3, edey1, rsreeni1, nroy\}@umbc.edu, archanm@smu.edu.sg 
}
}

\maketitle
\thispagestyle{plain}
\pagestyle{plain}

\begin{abstract}
In this demo paper, we design and prototype \paper~\cite{hasan2022rhythmedge}, a low-cost, deep-learning-based contact-less system for regular HR monitoring applications. \paper benefits over existing approaches by facilitating contact-less nature, real-time/offline operation, inexpensive and available sensing components, and computing devices. Our \paper system is portable and easily deployable for reliable HR estimation in moderately controlled indoor or outdoor environments. \paper measures HR via detecting changes in blood volume from facial videos (Remote Photoplethysmography; rPPG) and provides instant assessment using off-the-shelf commercially available resource-constrained edge platforms and video cameras. We demonstrate the scalability, flexibility, and compatibility of the \paper by deploying it on three resource-constrained platforms of differing architectures (NVIDIA Jetson Nano, Google Coral Development Board, Raspberry Pi) and three heterogeneous cameras of differing sensitivity, resolution, properties (web camera, action camera, and DSLR). \paper further stores longitudinal cardiovascular information and provides instant notification to the users. We thoroughly test the prototype stability, latency, and feasibility for three edge computing platforms by profiling their runtime, memory, and power usage. 

\end{abstract}


\section{Introduction}

Heart rate (HR) measures the number of heartbeats (contractions of the ventricles) per minute. HR can indicate various physiological conditions and is recommended to monitor regularly~\cite{zhang2016association}. \textit{\textbf{Wearable devices}} or \textit{\textbf{Video-based Remote Sensing}} leverage the Photoplethysmography (PPG) mechanism, an optical technique to detect volumetric changes in the peripheral blood circulation, to provide a non-invasive, low-cost regular HR measurement. Although the wearables are predominant as ubiquitous HR monitoring systems, they require close skin contact with specialized optical sensors~\cite{castaneda2018review}.

Alternatively, \textit{\textbf{Video-based Remote Sensing}} obtains remote Photoplethysmography (rPPG) in contact-less manner by capturing the variations in the light reflection off the human skin using off-the-shelf video sensors~\cite{verkruysse2008remote}. Due to the growing deployment of camera sensors, rPPG has the potential as a ubiquitous HR monitoring system while providing enhanced safety to healthcare officials and the general public. However, the low PPG SNR, variability in input videos~\cite{mcduff2017impact}, lack of robust PPG approximation model, camera sensor heterogeneity (sensor properties, resolution, sensitivity, frames per second (fps)), and the expensive video processing computation severely limit the rPPG applications. 

We address technical and scientific challenges towards developing an rPPG system for ubiquitous HR monitoring applications. In technical contributions, we posit appropriate settings to capture PPG using video cameras, facilitate off-line and real-time rPPG estimation, and devise systems compatible with various available low-cost resource constraint platforms and cameras. We also overcome the scientific challenges of developing a robust and efficient model to extract PPG for resource-efficient operation inside edge devices. 

\section{System Architecture}
\begin{figure}
    \centering
    \includegraphics[scale = .92]{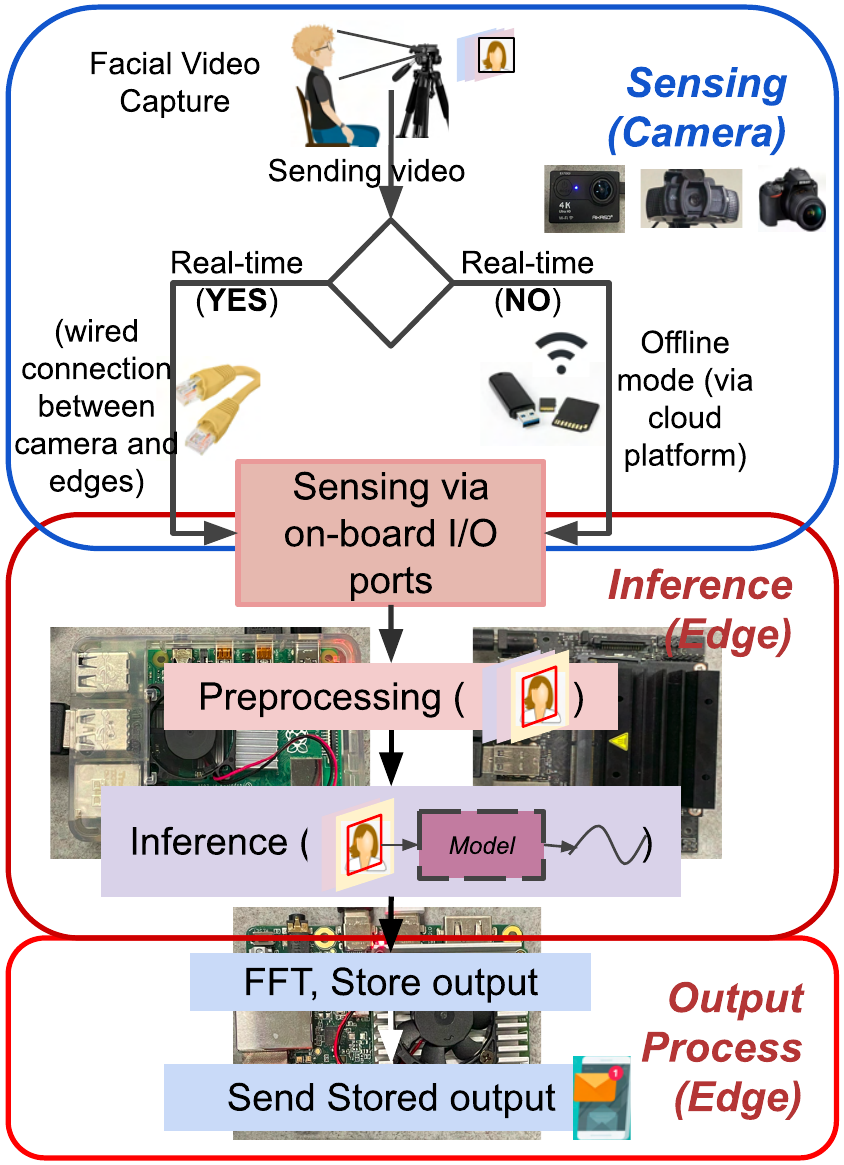}
    \caption{\paper system overview. The blue region and red region depict input sensing and edge device operations.}
    \label{fig:sys_protocol}
\end{figure}

The \paper system operates to infer PPG from video in real-time or offline mode. Both of the modes consist of three sequential steps: video input sensing, PPG inference from sensed videos, and Post-processing of inferred PPG [figure \ref{fig:sys_protocol}]. The input sensing step differs between the real-time and offline modes, while the later steps remain identical. 

\subsubsection{Input Sensing}

The edge devices receive the HD facial videos at $30$ fps for inferring PPG. During real-time operation, the edge devices capture the videos using their integrated cameras, whereas, in offline operation, the edge devices receive previously-stored videos via their input-output (I/O) ports. 

\textbf{Real-time Mode:} To facilitate real-time capturing of facial videos using edge devices, we integrate the USB-C compatible HD camera with their embedded USB-A port. We enable the Linux based \textit{ffmpeg} command to capture high-resolution $(1920 \times 1080 \times 3)$ video streams at $30$ fps using the device connected camera in real-time. Our re-programmable bash script controls and schedules the video recording based on application specifications. To get consistent quality input regardless of compatibility requirements, we employ the MJPEG compression scheme. \textbf{Offline Mode.} The edge devices receive previously captured facial videos (externally by DSLR, IR, and NIR cameras) via their on-board I/O communication system during the offline mode. The \paper can receive the input video using memory devices (Pen-drive, SD memory cards). Alternatively, we configure the devices to pull videos from cloud platforms (GitHub repository or google drive). Our automated Bash commands download the already uploaded videos from a pre-configured web-based cloud platform and store them in a specified directory for PPG inference.

\subsubsection{Data Preprocessing and Inference} The computing devices process the received video and apply the developed compressed model for PPG approximation. The model takes consecutive $40$ video frames of $100 \times 100$ resolution as a single input instance. Here, we address the technical challenges of incorporating video processing in a single pipeline under the memory, time, and power constraints. We install the python3 OpenCV library to perform video preprocessing. The OpenCV takes green channel of 1.33 second of video ($40$ green channel frames at $30$ fps), resize each frames to $100 \times 100$ to prepare one input instance of shape $1 \time 100 \times 100 \times 40$. In practice, OpenCV accumulates 22s of video ($660$ frames) and prepare 16 input instances $16 \times 100 \times 100 \times 40$ for infer at single model pass. Finally, the devices use the compressed tflite CamSense~\cite{hasan2022camsense} model and tensorflow runtime interpreter to extract PPG from processed video. In our prototype, the model approximate PPG of $64$ Hz, estimating approximately $85$ ($\frac{64 \times 40}{30} \approx 85$) points of PPG for $40$ frames and stack the consecutive outputs together.

\subsubsection{Output Processing} 
The computing devices further process the model-estimated PPG signal for HR approximation, store health data, and notify users. Utilizing the python Scipy library, we perform a Fast Fourier Transform (FFT) and select the FFT peak in the probable HR frequency range ($0.5$ to $4$ Hz) to extract HR from estimated PPG. The \paper can email the time-stamp and HR immediately to the user-specified email address via the Linux \textit{sendmail} commands. Further, the devices transfer the inferred PPG to the sever using onboard Wi-Fi to compute additional features, such as the peak to peak distance, output quality, and PPG irregularity. The server stores the users’ longitudinal PPG information with the date and time.

\section{Hardware and Software}

\subsubsection{Video Sensors}
In rPPG, the camera sensors need to capture high-fidelity skin light reflection at a high fps to embed PPG information. Additionally, the cameras should be compatible with the computing devices for a real-time operation. Based on the discussed criteria, we have selected three camera sensors: DSLR (offline but highest quality), Action Camera (AC), and Web-Camera (WC) [table \ref{tab:my-table1}].

\subsubsection{Limited Resource Computing Devices}
We have test \paper across different hardware architectures, and computing environments by considering three of the highly-used resource constrained devices: Coral Dev Board (GCDB), Jetson Nano (JN), and Raspberry Pi (RPi) [table \ref{tab:my-table2}].

\begin{table}[]
\centering
\caption{Overview of deployed camera sensors.}
\label{tab:my-table1}
\resizebox{\linewidth}{!}{%
\begin{tabular}{|l|lll|}
\hline
\multicolumn{1}{|c|}{\multirow{2}{*}{\textbf{Metrics}}} & \multicolumn{3}{c|}{\textbf{Camera Sensors}} \\ \cline{2-4} 
\multicolumn{1}{|c|}{} & \multicolumn{1}{l|}{\textbf{DSLR}} & \multicolumn{1}{l|}{\textbf{AC}} & \textbf{WC} \\ \hline
\textbf{Lens Property} & \multicolumn{1}{l|}{Flat Angle} & \multicolumn{1}{l|}{Wide Angle} & Flat Angle \\ \hline
\textbf{Compression} & \multicolumn{1}{l|}{H.264} & \multicolumn{1}{l|}{H.264} & H.264 \\ \hline
\textbf{Real time Mode} & \multicolumn{1}{l|}{No} & \multicolumn{1}{l|}{Yes} & Yes \\ \hline
\textbf{Bit rate} & \multicolumn{1}{l|}{10000} & \multicolumn{1}{l|}{2000} & 2000 \\ \hline
\textbf{FPS} & \multicolumn{1}{l|}{30} & \multicolumn{1}{l|}{30} & 30 \\ \hline
\textbf{WiFi} & \multicolumn{1}{l|}{No} & \multicolumn{1}{l|}{Yes} & No \\ \hline
\end{tabular}%
}
\end{table}

\begin{table*}[]
\centering
\caption{Overview of deployed edge devices.}
\label{tab:my-table2}
\resizebox{\textwidth}{!}{%
\begin{tabular}{|c|ccc|}
\hline
\multirow{2}{*}{\textbf{Metrics}} & \multicolumn{3}{c|}{\textbf{Computing Devices}} \\ \cline{2-4} 
 & \multicolumn{1}{c|}{\textbf{RPi}} & \multicolumn{1}{c|}{\textbf{JN}} & \textbf{GCDB} \\ \hline
\textbf{Processor} & \multicolumn{1}{c|}{\begin{tabular}[c]{@{}c@{}}Quad-core 1.5GHz Arm\\ Cortex-A72 based processor\end{tabular}} & \multicolumn{1}{c|}{\begin{tabular}[c]{@{}c@{}}Quad-core ARM Cortex-A57\\ 64-bit @ 1.42 GHz\end{tabular}} & Quad Cortex-A53, Cortex-M \\ \hline
\textbf{Operating System} & \multicolumn{1}{c|}{Raspbian Buster} & \multicolumn{1}{c|}{\begin{tabular}[c]{@{}c@{}}Linux4Tegra\\ (based on Ubuntu 18.04)\end{tabular}} & Mendel Linux \\ \hline
\textbf{Memrory} & \multicolumn{1}{c|}{2 GB LPDDR4-3200 SDRAM} & \multicolumn{1}{c|}{2-GB 64-bit LPDDR4 25.6 GB/s} & 1 GB LPDDR4 \\ \hline
\textbf{GPU} & \multicolumn{1}{c|}{--} & \multicolumn{1}{c|}{128 core Nvidia Maxwell} & Integrated GC7000 Lite Graphics \\ \hline
\textbf{Price (\$)} & \multicolumn{1}{c|}{100} & \multicolumn{1}{c|}{60} & 130 \\ \hline
\end{tabular}%
}
\end{table*}

\subsubsection{Device Configuration}
We install the necessary software packages with the peripherals in each device. Due to configuration disparity, we modify our installation strategy based on the selection set of computing devices and cameras. We have connected the cameras through the USB ports of the computing devices and captured the video while maintaining the framerate as desired using the \textit{ffmpeg} multimedia framework.
\begin{figure}[ht]
    \centering
    \includegraphics[scale = 0.6]{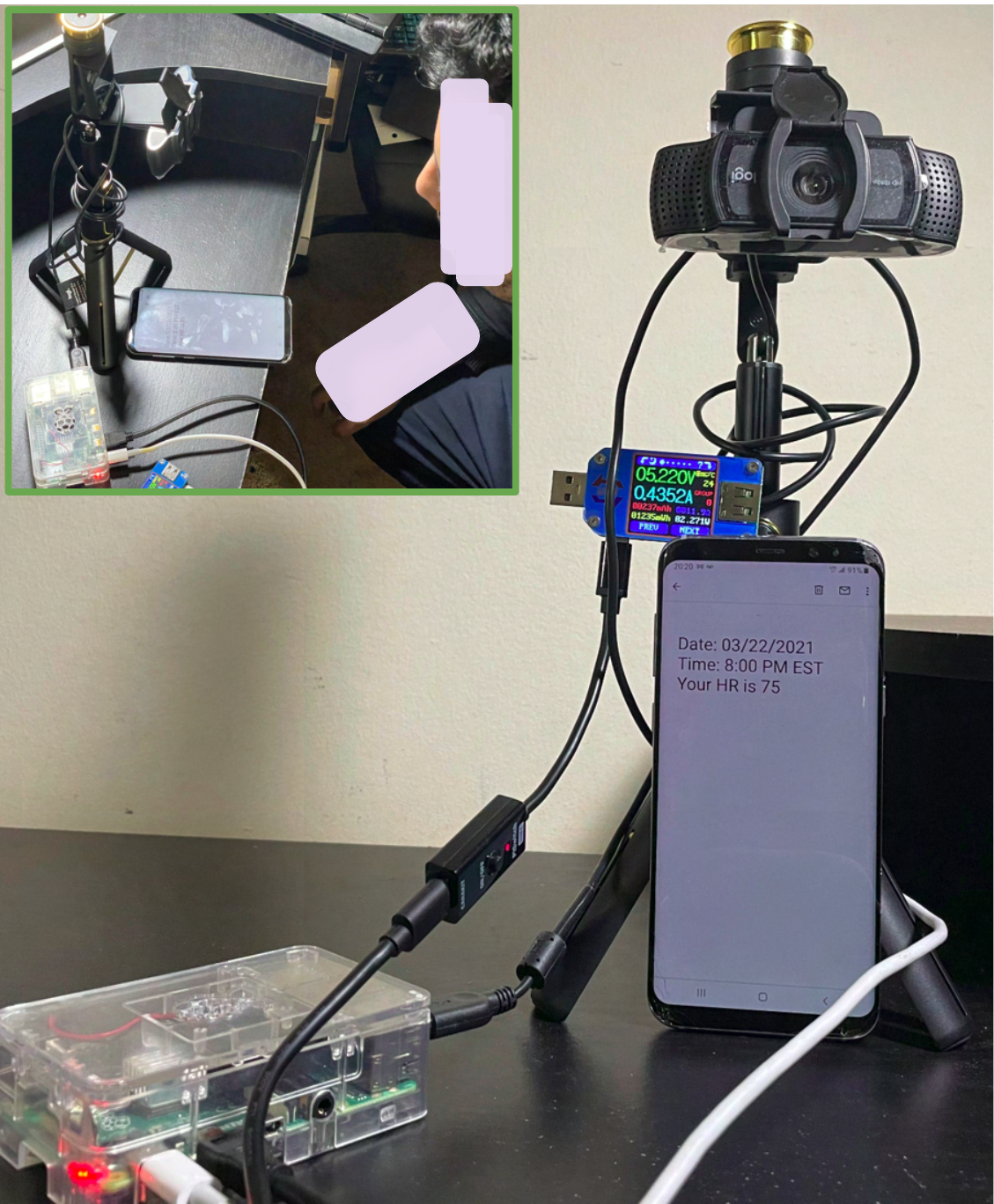}
    \caption{\paper demonstration in real-time}
    \label{fig:demo_fig}
\end{figure}

\section{Demo Description}
\subsubsection{Development Stage}
The \paper hardware consists of physical connections among three different modalities and wireless connections. To maintain the versatility of our rPPG system, we have experimented with three separate resource-constrained devices with different configurations. We chose one of those three computing devices and selected one of the two cameras (AC or WC) for real-time mode. We adjust minor settings in our bash file based on the chosen computing device and camera. Further, we attach a USB power meter in series with the power source to measure the power consumption. We have used the \textit{sendmail} protocol and $802.11$ac wireless network standard to transfer the messages. We have tested our rPPG system on 10 participants and achieved satisfactory performance in terms of accuracy and robustness.

\subsubsection{Demo Showcase plan}
To showcase the demo, the \paper prototype requires video captured under proper illumination (indoor or moderate sunlight) from stable, face-focused camera sensors from 3-6ft to the subjects [figure \ref{fig:demo_fig}]. The participant's email address may require to be included in our bash file to enable instant HR notification via email.

\subsubsection{Analysis on Computational Resource Usage}
Here, we discuss the protocol \paper system evaluation by measuring three relevant system parameters: Memory usage, Power consumption, and Latency. We have utilized Linux commands and Python libraries and deployed a USB power meter to track memory usage, execution time, and power consumption during the rPPG components' execution.

\textbf{Memory.}
We monitor the memory resources utilized during three rPPG tasks: (1) Input video processing with OpenCV, (2) Data processing for inference, and (3) Inferences. JN takes significantly more memory due to the GPU's lack of dedicated memory. The RPi and GCDB (slightly worse than RPi) are the more memory-efficient since they can reduce the memory demand of repetitive model loading by saving the model in the network-on-chip.
\textbf{Power.}
As part of our feasibility research, we profile the maximum and average power for different \paper components in all three devices. We measure the difference between idle and execution power multiple times and report their average. Among the rPPG components, capturing video consumes the most energy as the devices concurrently power the camera sensor and the computing device. For other components, we measure their power for processing $22s$ of video. 
\textbf{Execution Time.}
On three experimental resource-constrained platforms, we measure the execution time of the five rPPG system components: (1) OpenCV video processing, (2) video preprocessing, (3) Data loading, (4) Model load, and (5) Inference time and observe that the OpenCV processing time dominates across all computing machines, as it operates on high-definition (HD) and high fps video frames.

\section{Conclusion}
In this research, we have successfully developed and prototyped a real-time and offline rPPG system, \textit{RhythmEdge}, inside three limited resource devices by addressing relevant issues. We test and benchmark \paper by profiling the computation time, memory usage, and power consumption of the devices during the execution of rPPG system components. Our results demonstrate the feasibility of a real-time HR monitoring system inside ubiquitous computing devices. We demonstrate a low-cost, user-friendly HR monitoring prototype and envision its' mass application in the future.

\section*{Acknowledgements}
This research is supported by NSF CAREER grant 1750936, U.S. Army grant W911NF2120076. 


\bibliographystyle{ieeetr}
\bibliography{main}

\begin{thebibliography}{1}

\bibitem{hasan2022rhythmedge}
Z.~Hasan, E.~Dey, S.~R. Ramamurthy, N.~Roy, and A.~Misra, ``Rhythmedge:
  Enabling contactless heart rate estimation on the edge.'' Accepted for
  publication in 2022 IEEE International Conference on Smart Computing
  (SMARTCOMP), 2022.

\bibitem{zhang2016association}
D.~Zhang and et~al., ``Association between resting heart rate and coronary
  artery disease, stroke, sudden death and noncardiovascular diseases: a
  meta-analysis,'' {\em Cmaj}, vol.~188, no.~15, pp.~E384--E392, 2016.

\bibitem{castaneda2018review}
D.~Castaneda, A.~Esparza, M.~Ghamari, C.~Soltanpur, and H.~Nazeran, ``A review
  on wearable photoplethysmography sensors and their potential future
  applications in health care,'' {\em International journal of biosensors \&
  bioelectronics}, vol.~4, no.~4, p.~195, 2018.

\bibitem{verkruysse2008remote}
W.~Verkruysse, L.~O. Svaasand, and J.~S. Nelson, ``Remote plethysmographic
  imaging using ambient light.,'' {\em Optics express}, vol.~16, no.~26,
  pp.~21434--21445, 2008.

\bibitem{mcduff2017impact}
D.~J. McDuff, E.~B. Blackford, and J.~R. Estepp, ``The impact of video
  compression on remote cardiac pulse measurement using imaging
  photoplethysmography,'' in {\em 2017 12th IEEE International Conference on
  Automatic Face \& Gesture Recognition (FG 2017)}, IEEE, 2017.

\bibitem{hasan2022camsense}
Z.~Hasan, S.~R. Ramamurthy, and N.~Roy, ``Camsense: A camera-based contact-less
  heart activity monitoring,'' {\em Smart Health}, vol.~23, p.~100240, 2022.

\end{thebibliography}

\end{document}